\documentclass[10pt,twocolumn,letterpaper]{article}

\usepackage{iccv}
\usepackage{times}
\usepackage{epsfig}
\usepackage{graphicx}
\usepackage{amsmath}
\usepackage{amssymb}
\usepackage{graphicx}
\usepackage{amsmath}
\usepackage{amssymb}
\usepackage{booktabs}
\usepackage{float}
\usepackage{bbm}
\usepackage[table]{xcolor}
\usepackage{colortbl}
\usepackage{multirow}
\usepackage{mmstyles}
\usepackage{multicol}
\usepackage{bbm}
\usepackage{algorithm}
\usepackage{algpseudocode}
\definecolor{myy}{RGB}{126,95,0}
\definecolor{mygray}{gray}{.9}
\definecolor{mygray2}{gray}{0.5}
\definecolor{bblue}{RGB}{30,80,120}
\definecolor{mygray1}{gray}{.7}
\definecolor{ggray}{RGB}{127,127,127}
\definecolor{mygreen}{RGB}{93,174,86}
\definecolor{darkergreen}{RGB}{21, 152, 56}
\definecolor{red2}{RGB}{252, 54, 65}
\definecolor{y}{RGB}{255,230,0}
\definecolor{p}{RGB}{236,185,255}
\definecolor{g}{RGB}{0,235,0}
\usepackage{hhline}
\usepackage{booktabs}
\usepackage{algorithm}
\usepackage{algpseudocode}
\usepackage{subcaption}
\usepackage{pifont}
\usepackage{amsfonts}
\usepackage{amsmath}
\usepackage{arydshln}
\usepackage{multirow}
\usepackage{multicol}
\usepackage{xcolor}
\usepackage[export]{adjustbox}
\usepackage[normalem]{ulem}
\usepackage{xspace}
\usepackage{rotating}
\usepackage{xhfill}
\usepackage{ulem}
\usepackage{CJKulem}
\usepackage{soul}
\usepackage{color}
\definecolor{citecolor}{HTML}{229954}
\usepackage[pagebackref=true,breaklinks=true,colorlinks,citecolor=citecolor,bookmarks=false]{hyperref}

\def\iccvPaperID{3593} % *** Enter the ICCV Paper ID here

\iccvfinalcopy 
\ificcvfinal\fi

\begin{document}

%%%%%%%%% TITLE
\title{BoxDiff: Text-to-Image Synthesis with Training-Free Box-Constrained Diffusion}

\author{Jinheng Xie$^{1}$\quad Yuexiang Li$^{2*}$\quad Yawen Huang$^2$\quad Haozhe Liu$^{2,3}$\quad Wentian Zhang$^2$\\ Yefeng Zheng$^2$\quad Mike Zheng Shou$^{1*}$ \\
$^1$ Show Lab, National University of Singapore\quad  
$^2$ Jarvis Lab, Tencent \\
$^3$ AI Initiative, King Abdullah University of Science and Technology \\
{\tt\small \{sierkinhane,mike.zheng.shou\}@gmail.com}
% \\ \vspace{-1.2em}
\\
\tt\small\url{https://github.com/showlab/BoxDiff}
}

\twocolumn[{%
\maketitle

\renewcommand\twocolumn[1][]{#1}%
\begin{center}
    \centering
    \vspace{-22pt}
    \includegraphics[width=\linewidth]{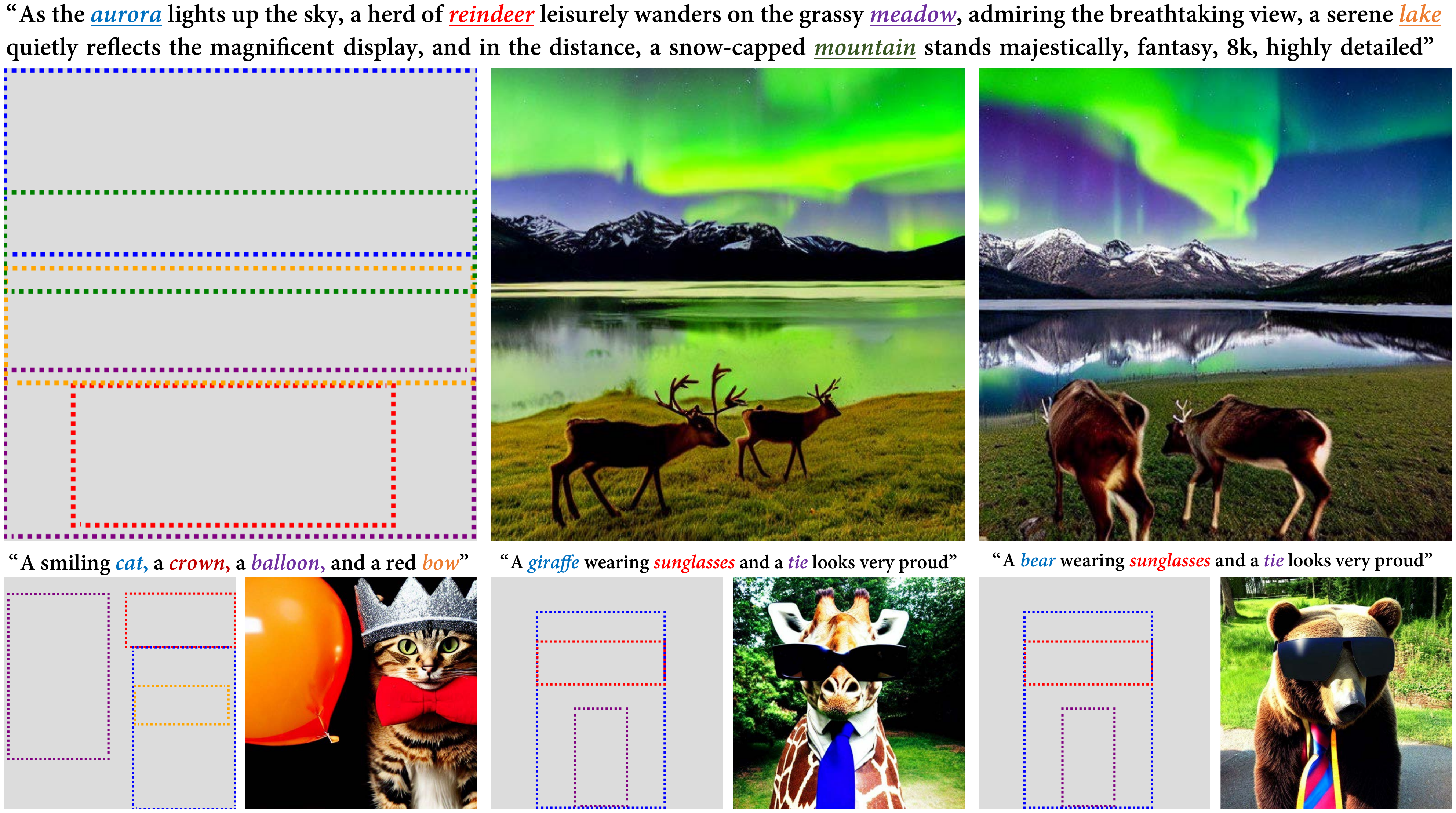}
    \vspace{-20pt}
    \captionof{figure}{In a training-free manner, BoxDiff consumes the simplest form of user-provided conditions, such as box or scribble, to control the location and scale of contents in the image synthesized by the pre-trained text-to-image diffusion model.}
    \label{fig:teaser}
    \vspace{-5pt}
\end{center}%
}]
\ificcvfinal\thispagestyle{empty}\fi
\let\thefootnote\relax\footnotetext{* Corresponding Author}
%%%%%%%%% ABSTRACT
\begin{abstract}
    Recent text-to-image diffusion models have demonstrated an astonishing capacity to generate high-quality images. However, researchers mainly studied the way of synthesizing images with only text prompts. While some works have explored using other modalities as conditions, considerable paired data, e.g., box/mask-image pairs, and fine-tuning time are required for nurturing models. As such paired data is time-consuming and labor-intensive to acquire and restricted to a closed set, this potentially becomes the bottleneck for applications in an open world. This paper focuses on the simplest form of user-provided conditions, e.g., box or scribble. To mitigate the aforementioned problem, we propose a training-free method to control objects and contexts in the synthesized images adhering to the given spatial conditions. Specifically, three spatial constraints, i.e., Inner-Box, Outer-Box, and Corner Constraints, are designed and seamlessly integrated into the denoising step of diffusion models, requiring no additional training and massive annotated layout data. Extensive experimental results demonstrate that the proposed constraints can control what and where to present in the images while retaining the ability of Diffusion models to synthesize with high fidelity and diverse concept coverage. 
\end{abstract}

%%%%%%%%% BODY TEXT
\section{Introduction}
Due to the large-scale publicly available image-text paired data from websites, recent text conditional auto-regressive and diffusion models, such as DALL-E 1 \& 2~\cite{dalle,dall2}, Imagen~\cite{imagen}, and Stable Diffusion~\cite{sd}, have demonstrated as one of the panaceas in generating images with high fidelity and diverse concept coverage. The excellent capacity of image synthesis increases the potential of these models for practical applications, \emph{e.g.,} art creation. However, most existing models can only be conditioned on class labels or text prompts. A few studies tried to use other modalities as conditions, \emph{e.g.,} spatial conditions, to further control the object or context synthesis. More fine-grained control on the location or scale of synthesized objects or contexts would widen the applications of text conditional generative models for the realistic scenario. For example, users can interactively design objects or contexts for human-in-the-loop art creation with additional spatial conditioning input. As a more user-friendly solution, this interactive cooperation with artificial intelligence (AI) would stimulate more potential for content creation.

Layout-to-image literature~\cite{li2021image, sun2021learning,sylvain2021object,yang2022modeling,zhao2019image,gafni2022make,balaji2022ediffi,avrahami2022spatext} has studied on the way to synthesize images adhering to the spatial conditioning input. However, the setting of these studies is restricted to the limited closed-set categories, which is infeasible to novel categories in open-world situations. Moreover, the previous studies followed the fully-supervised learning pipeline; hence, considerable paired box/skeleton/mask-image data is required for high-quality training. Since pixel-level annotation is time-consuming and labor-intensive to acquire, label efficiency gradually becomes the bottleneck of fully-supervised layout-to-image methods. Beyond text prompts as conditions, Stable Diffusion~\cite{sd} and ControlNet~\cite{controlnet} have also studied other modalities as conditioning input and provided qualitative results. In contrast to closed-set layout-to-image synthesis methods, Stable Diffusion, and ControlNet nurtured from large-scale image-text pairs have a strong perception of diverse visual concepts, \emph{e.g.,} different kinds of objects and contexts. Nevertheless, they also follow the general pipeline of layout-to-image literature in a fully-supervised manner, in which massive paired image-layout data is indispensable for high-quality training. Besides, the training period is time-consuming for train-from-scratch or fine-tuning.

\begin{figure}[t]
    \centering
    \includegraphics[width=\linewidth]{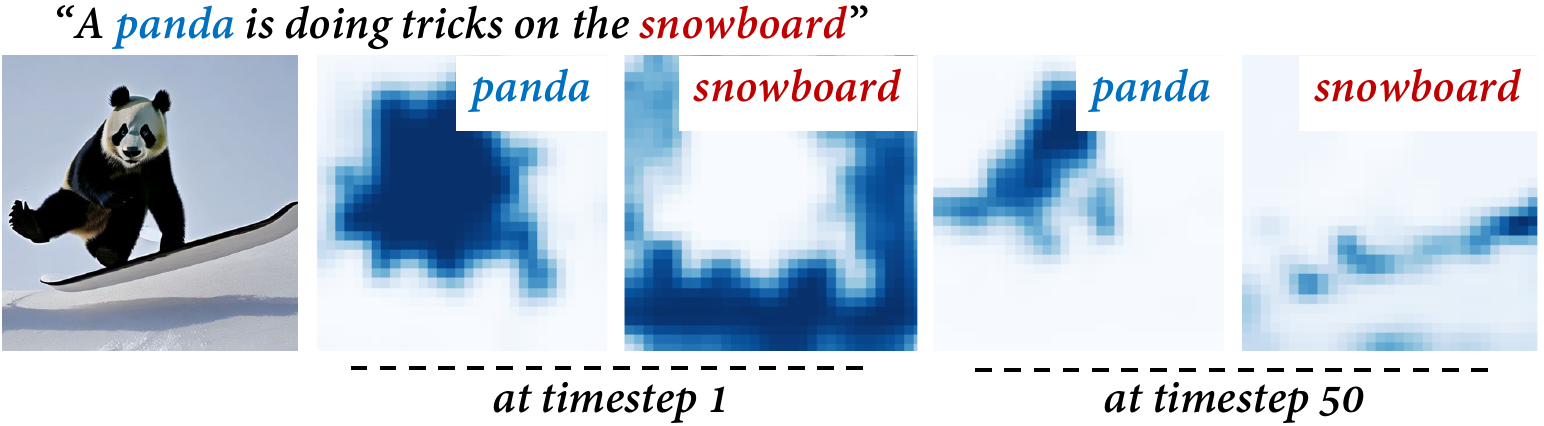}
    \vspace{-20pt}
    \caption{Cross-attentions between target text tokens, \emph{e.g.,} panda, bamboo, snowboard, and intermediate features of the denoiser, \emph{i.e.,} a UNet, in the Stable Diffusion model.} 
    \label{fig:cross_attn}
 \vspace{-15pt}
\end{figure}

In this paper, we focus on the most efficient setting for conditional image synthesis. Specifically, the simplest spatial conditions (or termed constraints), \emph{e.g.,} \textbf{box or scribble}, from users are adopted to seamlessly control object and context synthesis during the denoising step of Stable Diffusion models, \textbf{requiring no additional model training on the substantial paired layout-image data}. As shown in~\cite{p2p}, conditioning mechanisms incorporated in Stable Diffusion provide explicit cross-attentions between the given text prompt and intermediate features of the denoiser. Specific spatial attention maps for objects or contexts in the text prompt can be accordingly extracted. As the cross-attentions shown in Fig.~\ref{fig:cross_attn}, the spatial location of high-response attention, \emph{i.e.,} panda and snowboard, is perceptually equivalent to that of objects or contexts in the synthesized images. Hence, a simple idea to control the spatial location and scale of objects/contexts to be synthesized is adding guidance or constraints on the extracted cross-attentions. To achieve this goal, we propose a \textbf{training-free} approach, namely Box-Constrained Diffusion (BoxDiff), by adding three spatial constraints, \emph{i.e.,} \underline{Inner-Box, Outer-Box, and Corner Constraints}, on the cross-attentions extracted at each denoising timestep. This plays a role in pointing out directions to update the noised latent vector, which consequently leads synthesized objects or contexts to gradually follow the given spatial conditions. Furthermore, since strong constraints applied to the cross-attentions will affect the denoising step of diffusion models, impairing the fidelity of the resulting synthesized images, we also explore a manner of representative sampling to mitigate the problem. Samples synthesized by the proposed BoxDiff can be found in Fig.~\ref{fig:teaser}.

The main contributions of this paper are summarized as:
\begin{itemize}
    \item We propose a training-free approach, termed Box-Constrained Diffusion (BoxDiff), for text-to-image synthesis following the given spatial conditions, requiring no additional model training and massive paired layout-image data.
    
    \item The proposed spatial constraints can be seamlessly incorporated into the denoising step, which retains the strong perception of diverse visual concepts of Stable Diffusion. Hence, our method can synthesize various novel objects and contexts beyond the closed world.

    \item Extensive experiments demonstrate that the proposed training-free BoxDiff can synthesize photorealistic images following the given spatial conditions. 

\end{itemize}

\section{Related Work}
\textbf{Diffusion Models:} Recently, diffusion models have ushered in a new era of image generation. It consists of a forward process, \emph{i.e.,} adding noise, and a reverse process, \emph{i.e.,} removing noise. The denoising diffusion probabilistic model (DDPM)~\cite{sohl2015deep,ho2020denoising} learns to invert a parameterized Markovian image noising process. Given isotropic Gaussian noise samples, they can transform them into signals, \emph{e.g.,} images, by iteratively removing the noise. Beyond pure noise to image fashion, class-conditional and image-guided synthesis have also been explored~\cite{choi2021ilvr,dhariwal2021diffusion,meng2021sdedit}. In contrast to denoising in the pixel space, Rombach \etal~\cite{sd} proposed to operate on the compressed latent space by employing an autoencoder. This significantly lowers the training costs and speeds up the inference time while retaining the ability to generate high-quality images.

\textbf{Text-to-Image Models:} Recently, large-scale image-text pairs available on the Internet dramatically enabled generative models, \emph{e.g.,} DALL-E~\cite{dalle}, Imagen~\cite{imagen}, and Stable Diffusion~\cite{sd}, to synthesize images in higher quality and richer diversity. \cite{feng2022training} and recently introduced \cite{attend} operated on the cross-attention for better content consistency to subjects in text prompts. However, they are chiefly conditioned on the text prompts or class labels. As compensation, a few works~\cite{gafni2022make,avrahami2022spatext,balaji2022ediffi} have been proposed to handle additional spatial layout conditions, but most of them require additional model training or have only a limited scope of knowledge. For example, Gafni \etal~\cite{gafni2022make} incorporated semantic maps to control objects \& contexts in the generated images. However, it is restricted to the closed-set world (only 158 categories). Concurrently, Balaji \etal~\cite{balaji2022ediffi} illustrated that a modification of the attention map can lead to corresponding changes in the synthesized objects. Motivated by the above, we are interested in controlling object synthesis by the simplest form of conditions, \emph{e.g.,} box or scribble, from users, potentially motivating simpler and more efficient interactive cooperation of image synthesis.

\textbf{Layout-to-Images Models:} Traditional layout-to-image literature~\cite{li2021image, sun2021learning,sylvain2021object,yang2022modeling,zhao2019image,gafni2022make} has been focused on how to synthesize images adhering to the given bounding boxes of object categories. Generally, they follow the pipeline, \emph{i.e.,} training and validation, to obtain layout-to-image models, and promising results have been obtained. However, they are trapped in a dilemma of time-consuming and labor-intensive annotation like box/mask-image paired data. In addition, they are greatly restricted to a fixed number of categories, failing to synthesize novel categories in the open world. The image quality of such models is also lower than that of the recently introduced large-scale image-text-pairs-driven generative models. Recently, fine-tuning Stable Diffusion models to adhere to additional layout information has also been explored in \cite{gligen} and \cite{yang2023reco}. Compared to the above methods, we propose a training-free approach by adding the simplest constraints, \emph{e.g.,} box or scribble, from users to the denoising step of Stable Diffusion models. It requires no additional training and paired layout-image data. Besides, the proposed approach has the ability to synthesize a wide range of visual concepts rather than the limited closed set. Note that, there are many concurrent works~\cite{phung2023grounded, chen2023training, bansal2023universal, ma2023directed} that studied a similar area. For example, both Chen \etal~\cite{chen2023training} and Phung \etal~\cite{phung2023grounded} operated constraints on the cross-attention to control the synthetic contents. More recently, VisorGPT~\cite{xie2023visorgpt}, LMD~\cite{lian2023llmgrounded}, LayoutGPT~\cite{feng2023layoutgpt}, and Control-GPT~\cite{zhang2023controllable} have been proposed to plan visual layouts for image synthesis models. Along with image personalization models such as DreamBooth~\cite{ruiz2023dreambooth}, Textual Inversion~\cite{gal2022image}, Mix-of-Show~\cite{gu2023mixofshow}, and Perfusion~\cite{tewel2023key}, our BoxDiff has become increasingly feasible to create a more complex scene with personalized contents from Diffusion Models.

\begin{figure*}[t]
    \centering
    \includegraphics[width=\linewidth]{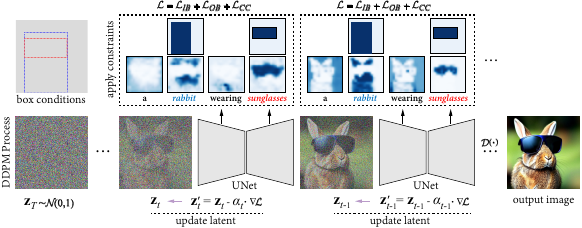}
    \vspace{-20pt}
    \caption{Overview of our BoxDiff. Given the box conditions, we transform them into a set of binary spatial masks. A latent $\vz_T$ sampled from Normal distribution $\mathcal{N}(0,1)$ is passed to the denoiser, \emph{i.e.,} UNet, to obtain the denoised latent. At timestep $t$, $\vz_t$ is first passed to UNet to get the cross-attention maps, on which the proposed constraints, \emph{i.e.,} $\mathcal{L} = \mathcal{L}_{IB} + \mathcal{L}_{OB} + \mathcal{L}_{CC}$, are applied. Subsequently, the current latent $\vz_t$ can be updated by the gradient $\nabla \mathcal{L}$ to get $\vz^{\prime}_t$ for the denoising step.} 
    \label{fig:method}
 \vspace{-15pt}
\end{figure*}

\section{Preliminaries: Stable Diffusion}
Different from~\cite{ho2020denoising,dhariwal2021diffusion}, the Stable Diffusion model efficiently operates on the latent space. Specifically, an autoencoder consisting of an encoder $\mathcal{E}$ and decoder $\mathcal{D}$ is trained with a reconstruction objective. Given an image $\vx$, the encoder $\mathcal{E}$ maps it to a latent $\vz$, and the decoder $\mathcal{D}$ reconstructs the image from the latent, \emph{i.e.,} $\widetilde{\vx}=\mathcal{D}(\vz)=\mathcal{D}(\mathcal{E}(\vx))$. In this way, at each timestep $t$, a noisy latent $\vz_t$ can be obtained. Beyond the routine training scheme, Stable Diffusion devises conditioning mechanisms to control the synthesized image content by an additional input, \emph{e.g.,} a text prompt $\vy$. The text prompt is first pre-processed to text tokens $\tau_{\theta}(\vy)$ by the text encoder of pre-trained CLIP~\cite{clip}. The DDPM model $\epsilon_{\theta}$ can then be trained via:
\begin{equation}
    \mathcal{L}_{DDPM} = \mathbb{E}_{\vz\sim \mathcal{E}(\vx),\vy,\epsilon \sim \mathcal{N}(0,1),t}\left[||\epsilon - \epsilon_{\theta}(\vz_t,t,\tau_{\theta}(\vy))||^2_2\right],
\end{equation}
where UNet~\cite{unet} enhanced with self-attention and cross-attention layers is adopted as the denoiser $\epsilon_{\theta}$. During training, given a noised latent $\vz_t$ at timestep $t$ and text tokens $\tau_{\theta}(\vy)$, denoiser $\epsilon_{\theta}$ is tasked with predicting the noise $\epsilon$ added to the current latent.

In inference, a latent $\vz_T$ is sampled from the standard normal distribution $\mathcal{N}(0,1)$ and the DDPM is used to iteratively remove the noise in $\vz_T$ to produce $\vz_0$. In the end, the latent $\vz_0$ is passed to the decoder $\mathcal{D}$ to generate an image $\widetilde{\vx}$.

\section{Methodology}
In this section, we present the proposed BoxDiff approach and spatial constraints in detail.

\subsection{Cross-Modal Attention}
Conditioning mechanisms in the Stable Diffusion model can explicitly form the cross-attentions between text tokens and intermediate features of the denoiser $\epsilon_{\theta}$. In the denoising step, given the conditioning text tokens $\tau_{\theta}(\vy)$ and intermediate features $\varphi(\vx_t)$, the cross-attention $\mA$ can be accordingly acquired:
\begin{equation}
    \mA = \text{Softmax}(\mQ\mK^{\top} / \sqrt{d}),
\end{equation}
\begin{equation}
    \mQ=\mW_Q \varphi(\vx_t),\; \mK=\mW_K \tau_{\theta}(\vy),
\end{equation}
where $\mQ, \mK$ are the projection of intermediate features $\varphi(\vx_t)$ and text tokens $\tau_{\theta}(\vy)$ by two learnable matrices $\mW_Q, \mW_K$, respectively. At each timestep $t$, given $\tau_{\theta}(\vy)$ with $N$ text tokens $\{\mathbf{s}_1,\cdots,\mathbf{s}_N\}$, the cross-attention $\mA^t$ containing $N$ spatial attention maps $\{\mA_1^t,\cdots,\mA_N^t\}$ can be consequently obtained. Here, following~\cite{attend}, we remove the cross-attention between the start-of-text token (\emph{i.e.,} \text{[sot]}) and intermediate features before applying Softmax$(\cdot)$. Besides, a Gaussian filter is applied to smooth the cross-attentions along the spatial dimension. Therefore, the aforementioned operations bring an enhancement on cross-attentions between actual subject tokens, \emph{e.g.,} object or context, with the intermediate features. Cross-attention can be performed at different scales, \emph{i.e.,} $64\times 64, 32\times 32, 16\times 16,\text{and } 8\times 8$. Following~\cite{p2p}, we operate the proposed constraints on the cross-attentions with a resolution of $16\times 16$ as the inherent sufficient semantic information. 

\subsection{Box-Constrained Diffusion}
Given a text prompt with a set of target tokens $\mathcal{S} = \{\mathbf{s}_i\}$ and a set of user-provided object or context locations $\mathcal{B} = \{\vb_i\}$ as spatial conditions, a set of corresponding spatial cross-attention maps $\mathcal{A}^{t} = \{\mA^{t}_i\}$ between target tokens and intermediate features can be accordingly obtained. For example, cross-attention over target tokens such as ``rabbit'' and ``sunglasses'' can be yielded (as shown in Fig.~\ref{fig:method}). Each location $\vb_i$ contains the user-provided top-left and bottom-right coordinates $\{(x^i_1,y^i_1),(x^i_2,y^i_2)\}$.

It can be observed from Fig.~\ref{fig:cross_attn} that, during the denoising step of the Stable Diffusion model, the location and scale of high response regions in the cross-attention map are perceptually equivalent to that of synthesized objects in the decoded image $\widetilde{\vx}$. This motivates us that constraints can be added on the cross-attention to control the synthesis of target objects in the image $\widetilde{\vx}$. As shown in Fig.~\ref{fig:method}, given user-provided location, \emph{i.e.,} $\{\vb_i\}$, a set of binary spatial masks $\mathcal{M}=\{\mM_i\}$ can be transformed from the top-left and bottom-right coordinates, where each $\mM_i \in \mathbb{R}^{16\times 16}$. Our target is to synthesize target objects approaching the mask regions. To achieve this goal, we propose three spatial constraints, \emph{i.e.,} Inner-Box, Outer-Box, and Corner Constraints, over the target cross-attention maps $\mathcal{A}^t$ to gradually update the latent $\vz_t$ such that the location and scale of synthesized objects will be consistent with the mask region. Henceforward, the diffusion model with our three constraints is named as Box-Constrained Diffusion (BoxDiff).

\textbf{Inner-Box Constraint:} To ensure the synthesized objects will approach the user-provided locations, a simple solution is to ensure that high responses of cross-attention are only in the mask regions. To this end, we propose the inner-box constraint as below:
\begin{equation}
\label{eq:l1}
   \mathcal{L}_{\mathbf{s}_i}^1 = 1 - \frac{1}{P}\sum \textbf{topk}\left(\mA^t_i \cdot \mM_i, P \right),
\end{equation}
\begin{equation}
   \mathcal{L}_{IB} = \sum_{s_i \in \mathcal{S}} \mathcal{L}_{\mathbf{s}_i}^1,
\end{equation}
    where $\textbf{topk}(\cdot, P)$ means that $P$ elements with the highest response would be selected. As observed in the experiments, constraints added on all elements in the cross-attention map potentially lead to a collapse of image fidelity. Besides, constraints on only a few elements with high responses are sufficient to affect the synthesis of objects, which can reduce the impact of constraints and prevent the failure of denoising. Hence, only $P$ elements are constrained to update the latent $\vz_t$. Binary mask $\mM_i$ in Eq.~(\ref{eq:l1}) aims to mask out elements of the cross-attention maps within the mask regions and $\mathcal{L}_{IB}$ plays a role to maximize the response of the mask-out elements. 

\textbf{Outer-Box Constraint:} However, the involvement of Inner-Box Constraint can only ensure that the user-provided regions in $\widetilde{\vx}$ will contain objects. It cannot guarantee that no object pixels are synthesized out of the user-provided boxes. To prevent the object from moving out of  the target regions, we propose the outer-box constraint as follows:
\begin{equation}
\label{eq:l2}
   \mathcal{L}_{\mathbf{s}_i}^2 = \frac{1}{P}\sum \textbf{topk}\left(\mA^t_i\cdot (1-\mM_i), P \right),
\end{equation}
\begin{equation}
   \mathcal{L}_{OB} = \sum_{s_i \in \mathcal{S}} \mathcal{L}_{\mathbf{s}_i}^2.
\end{equation}
In Eq.~(\ref{eq:l2}), we get the reversion of mask $(1-\mM_i)$ to mask out elements of the cross-attention map beyond the target regions. Here, $\mathcal{L}_{OB}$ aims to minimize the response of cross-attentions out of the target regions. Note that two constraints $\mathcal{L}_{IB}$ and $\mathcal{L}_{OB}$ work in a complementary manner. 

\begin{figure}[h]
    \centering
    \vspace{-10pt}
    \includegraphics[width=\linewidth]{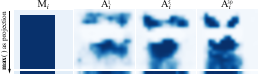}
    \vspace{-23pt}
    \caption{Examples of projection on the x-axis.} 
    \label{fig:fixed_loc}
 \vspace{-10pt}
\end{figure}

\textbf{Corner Constraint:} Since there are only weak spatial conditions, \emph{i.e.,} box or scribble, from users, exact boundary pixels of objects or contexts are not available to restrict the scale. Hence, there will be some tricky solutions, \emph{e.g.,} the target objects are synthesized on a smaller scale than the box regions, with only the above two constraints. In this regard, we propose the corner constraint at the projection of the x-axis and y-axis, respectively. First, we project each target mask $\mM_i$ and cross-attention $\mA^t_i$ on the x-axis via the \textbf{max} operation as below:
\begin{equation}
    \vm_x(k) = \textbf{max}_{j=1,\cdots,H} \{\mM_i(j,k)\}, 
\end{equation}
\begin{equation}
    \va^t_x(k) = \textbf{max}_{j=1,\cdots,H} \{\mA^t_i(j,k)\}, 
\end{equation}
where $\vm_x \in \mathbb{R}^{W}$ and $\va^t_x \in \mathbb{R}^{W}$. $\vm_x$ is employed as the target and we aim to optimize $\va^t_x$ close to $\vm_x$: 
\begin{equation}
   \label{eq:l3} 
   \mathcal{L}_{\mathbf{s}_i}^3 = \frac{1}{L}\sum\textbf{sample}\left(\{|\vm_x(k)-\va^t_x(k)|\}_{k=1}^W, L, x^i_1,x^i_2 \right),
\end{equation}
where $\textbf{sample}(\cdot, L, x^i_1,x^i_2)$ indicates a uniform sampling of $L$ error terms from the set $\{|\vm_x(k)-\va_x^t(k)|\}_{k=1}^W$ around the given corner coordinates $x^i_1$ and $x^i_2$ at x-axis. It plays the same role as the \textbf{topk} sampling in Eqs.~(\ref{eq:l1}) and (\ref{eq:l2}).

\begin{figure*}[t]
    \centering
    \includegraphics[width=\linewidth]{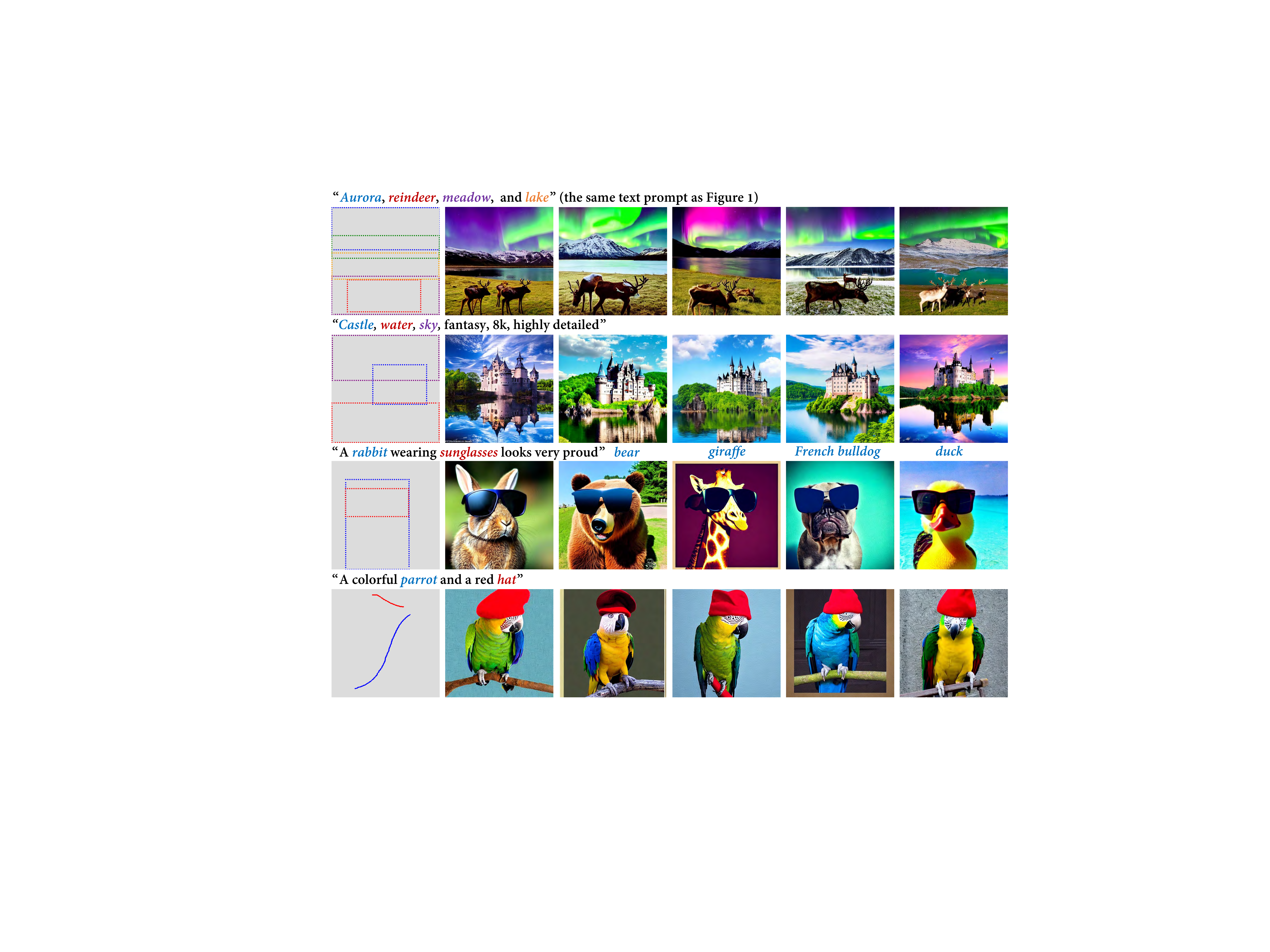}
    \vspace{-23pt}
    \caption{Multiple samples synthesized with fixed spatial conditioning inputs.} 
    \label{fig:fixed_loc}
 \vspace{-18pt}
\end{figure*}

For y-axis projection, same operations are performed:
\begin{equation}
    \vm_y(j) = \textbf{max}_{k=1,\cdots,W} \{\mM_i(j,k)\}, 
\end{equation}
\begin{equation}
    \va^t_y(j) = \textbf{max}_{k=1,\cdots,W} \{\mA^t_i(j,k)\}, 
\end{equation}
\begin{equation}
   \label{eq:l4} 
   \mathcal{L}_{\mathbf{s}_i}^4 = \frac{1}{L}\sum\textbf{sample}\left(\{|\vm_y(j)-\va_y^t(j)|\}_{j=1}^H, L, y^i_1, y^i_2 \right).
\end{equation}

The corner constraint is the summation of $\mathcal{L}_{\mathbf{s}_i}^3$ and $\mathcal{L}_{\mathbf{s}_i}^4$ as below:
\begin{equation}
    \mathcal{L}_{CC} = \sum_{s_i \in \mathcal{S}} \mathcal{L}_{\mathbf{s}_i}^3 + \mathcal{L}_{\mathbf{s}_i}^4.
\end{equation}

At each timestep, \textbf{overall constraints} is formulated as:
\begin{equation}
\label{eq:L}
    \mathcal{L} = \mathcal{L}_{IB} + \mathcal{L}_{OB} + \mathcal{L}_{CC}.
\end{equation}

Having computed the loss $\mathcal{L}$, the current latent $\vz_t$ can be updated with a step size of $\alpha_t$ as follow:
\begin{equation}
    \vz_{t}^{\prime} \leftarrow \vz_t - \alpha_t \cdot \nabla \mathcal{L},
\end{equation}
where $\alpha_t$ decays linearly at each timestep. With a combination of the aforementioned constraints, $\vz_t$ at each timestep gradually moves toward the direction of generating high-response attention in the given location and with a similar scale to the box, which leads to a synthesis of target objects in the user-provided box regions. 

\textbf{Note:} We prioritize enabling users to provide conditions in the possibly simplest way, \emph{i.e.,} bounding boxes. Beyond that, BoxDiff can interact with other types of conditions such as scribble. More details are in the appendix. Additionally, BoxDiff can be used as a plug-and-play component in many diffusion models, including GLIGEN~\cite{gligen}.

\section{Experiments}
\label{experiments}
\subsection{Experimental Setup}
\textbf{Datasets:} Current layout-to-image methods are mainly trained on paired layout-image data of COCO-Stuff~\cite{caesar2018coco} or VG~\cite{krishna2017visual}. It is unfair to directly make comparisons between our training-free BoxDiff and the fully-supervised methods. Hence, we propose to compare performance on a new dataset. Details can be found in the appendix. Specifically, we collect a set of images (no intersection with COCO and VG) and use YOLOv4~\cite{bochkovskiy2020yolov4} to detect objects. For evaluation, we consider two types of situations: i) a single instance, \emph{i.e.,} ``a \{\}"; ii) multiple instances, \emph{i.e.,} ``a \{\}, a \{\}". In this way, 189 different text prompts are combined with conditional boxes for image synthesis. 

% For layout-to-image methods, we map the collected categories to the order of COCO-Stuff for conditional image synthesis.

\textbf{Evaluation Metrics:} To validate the effectiveness of our BoxDiff, YOLOv4 is employed to detect object-bounding boxes and predict classification scores on the synthesized images. YOLO score~\cite{li2021image}, including AP, AP$_{50}$ and AP$_{75}$, is adopted to evaluate the precision of the conditional synthesis. Additionally, we employ a metric of Text-to-Image Similarity (T2I-Sim) to explicitly evaluate the correctness of semantics in the synthesized images. In particular, synthesized images and the text prompts, \emph{e.g.,} ``a photo of \{\}'' or ``a photo of \{\} and \{\}'', are passed to the image and text encoder of pre-trained CLIP~\cite{clip}, respectively, to calculate their similarity (\emph{i.e.,} T2I-Sim). In CLIP feature space, the similarity can reflect whether the semantics of objects or contexts are correctly presented in the images. 

\begin{table}[t]
    \centering
    \caption{Ablation studies on various components.}
        \vspace{-10pt}
    \resizebox{\linewidth}{!}{
        \begin{tabular}{l c c c c c c}
            \toprule[1.5pt]
            $\mathcal{L}_{IB}$ & $\mathcal{L}_{OB}$ & $\mathcal{L}_{CC}$ & $\textbf{sample}(\cdot)$ & $\textbf{topk}(\cdot)$ & T2I-Sim ($\uparrow$) &  AP ($\uparrow$)  \\
            \midrule
            \multicolumn{5}{c}{Stable Diffusion} & 0.3511 & 2.8 \\
            % \multicolumn{5}{c}{Composable Diffusion} \\
            % \multicolumn{5}{c}{Structure Diffusion} \\
            \midrule
            \checkmark  &   &   &  \checkmark  &\checkmark & 0.3516    & 9.8  \\
            \checkmark & \checkmark  &   &  \checkmark & \checkmark & 0.3518  & 20.2 \\
            \rowcolor{mygray}
            \checkmark & \checkmark & \checkmark  & \checkmark &  \checkmark & 0.3513  & 22.3 \\
            \checkmark & \checkmark & \checkmark  & \checkmark  & &  0.3489   & 24.8  \\
            \checkmark & \checkmark & \checkmark  &  & \checkmark  & 0.3472 &    7.7  \\
            
            \bottomrule[1pt] 
    \end{tabular}}
    \label{tab:ablation}
    \vspace{-25pt}
\end{table}

\subsection{Ablation Studies}
\begin{figure*}[t]
    \centering
    \includegraphics[width=\linewidth]{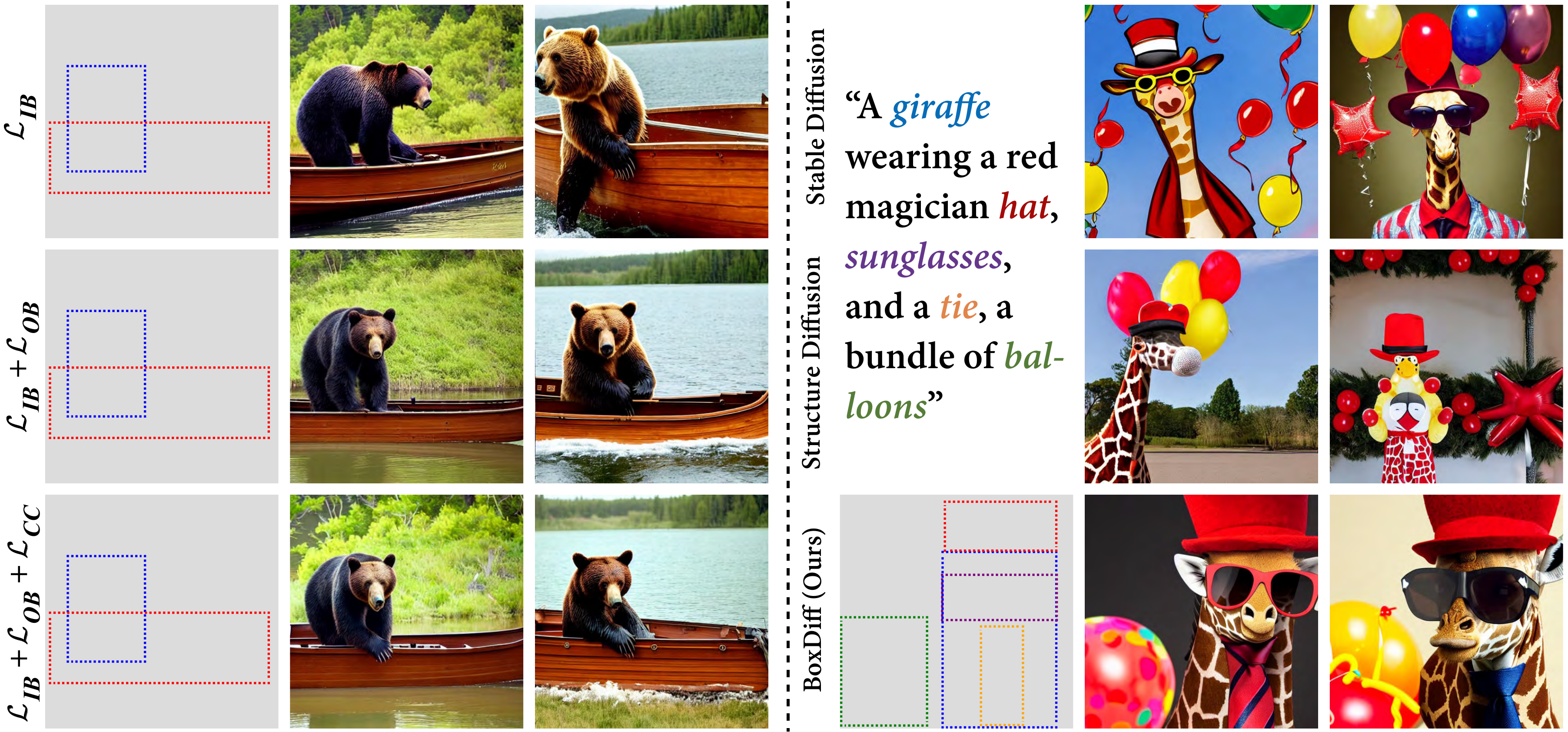}
    \vspace{-23pt}
    \caption{Left: Ablation studies on various combinations of constraints. Right: Visual comparison with \cite{sd} and \cite{feng2022training}.} 
    \label{fig:compar_and_ablation}
 \vspace{-18pt}
\end{figure*}

\textbf{Impact of Various Constraints:}
To validate the impact of $\mathcal{L}_{IB}$, $\mathcal{L}_{OB}$, and $\mathcal{L}_{CC}$, we perform ablation studies on different combinations of constraints, and the results are listed in Table~\ref{tab:ablation}. As shown, the model achieves a T2I-Sim of 0.3516 and an AP of 9.8 in terms of YOLO score with only the inner-box constraint $\mathcal{L}_{IB}$. Such a result reveals that the synthesized objects are mostly not consistent with the conditional spatial input. As $\mathcal{L}_{IB}$ and $\mathcal{L}_{OB}$ work complementary to restrict the cross-attention of objects inside the conditional boxes, a higher YOLO score of 20.2 AP is achieved on the synthesized images. When corner constraint $\mathcal{L}_{CC}$ is involved to limit the corner elements on the projection of cross-attention, the scales of synthesized objects are guaranteed to be consistent with the given bounding box conditions, which accordingly increases the AP from 20.2 to 22.3. Obviously, the proposed constraints are effective in controlling the location and scale of synthesized objects. 
Visual variations can be found on the left in Fig.~\ref{fig:compar_and_ablation}.

\textbf{Impact of Representative Sampling:} As aforementioned, adding constraints to all elements in the cross-attentions may potentially affect image synthesis. The quantitative evaluation is presented in Table~\ref{tab:ablation}. Without $\textbf{topk}(\cdot)$ in Eq.~(\ref{eq:l1}) and Eq.~(\ref{eq:l2}), though there is an improvement of AP, T2I-Sim of the synthesized images decreases. This accordingly represents that the consistency between semantics synthesized in the images and the given text prompts is impaired, and the image quality is decreased. When $\textbf{sample}(\cdot)$ in Eq.~(\ref{eq:l3}) and Eq.~(\ref{eq:l4}) is removed, T2I-Sim of synthesized images significantly degrades. The removal of $\textbf{sample}(\cdot)$ also impairs the consistency of synthesized objects to the conditional input, leading to a lower AP. Hence, we adopt $\textbf{topk}(\cdot)$ and $\textbf{sample}(\cdot)$ for the better image quality and consistency with the text prompts.

\textbf{Impact of sampling in $\mathcal{L}_{IB}$ and $\mathcal{L}_{OB}$.} In Table.~\ref{tab:various_sampling}. One can observe that while sampling all pixels in $\mathcal{L}_{IB}$ and $\mathcal{L}_{OB}$ can lead to a more precise synthesis adhering to the conditions, the quality of synthetic contents will be correspondingly degraded (lower T2I-Sim than that of \textbf{topk($\cdot$)}). Randomly sampling pixels cannot effectively maximize the activation of foreground pixels and may activate background regions in cross-attention, leading to significant degradation of the AP and T2I-Sim. To balance these trade-offs, we propose \textbf{topk($\cdot$)}, which achieves the best synthetic quality while maintaining a relatively good AP.

\begin{figure*}[t]
    \centering
    \includegraphics[width=\linewidth]{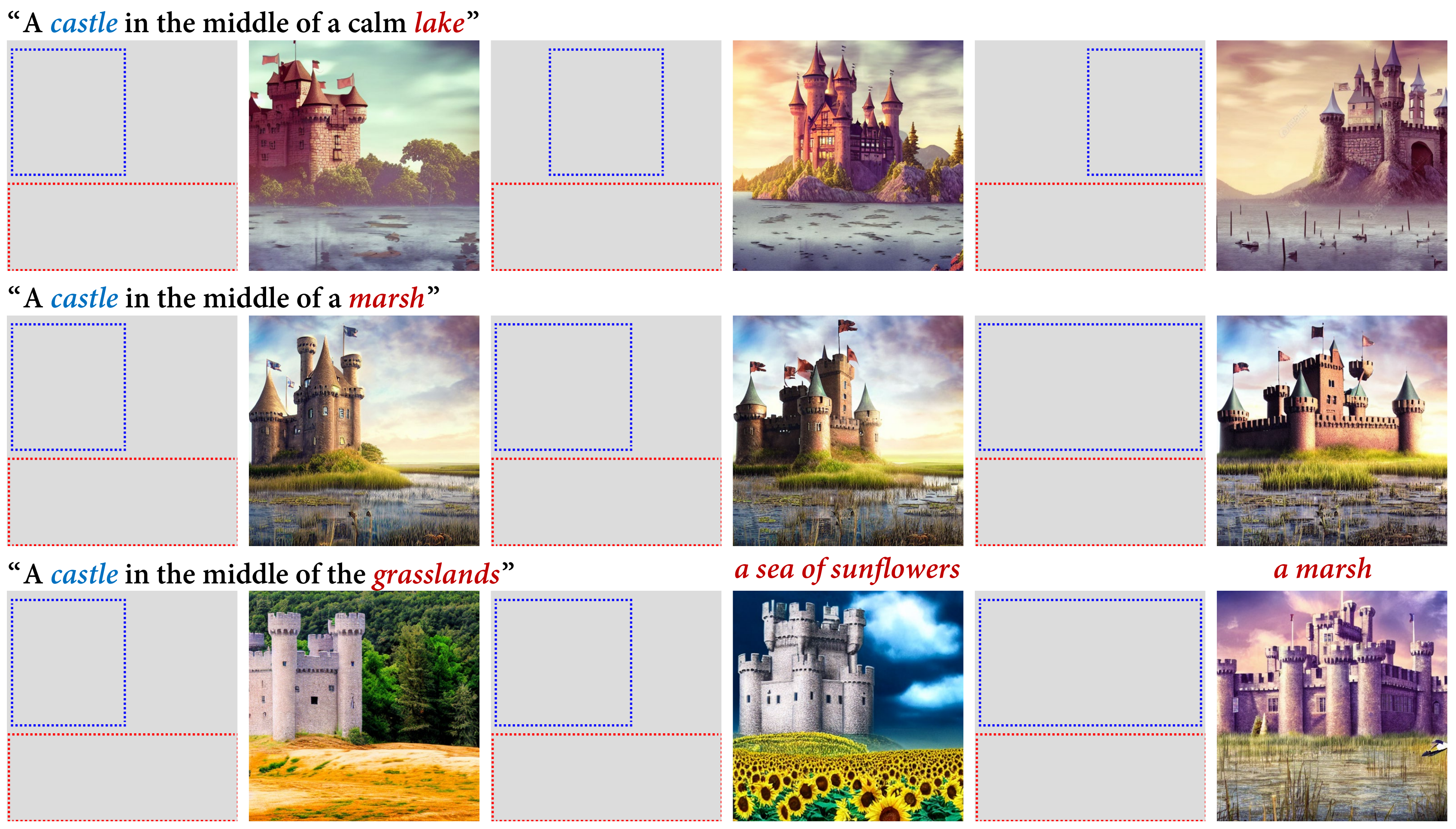}
    \vspace{-23pt}
    \caption{Synthesized samples obtained with various spatial conditioning inputs, \emph{e.g.,} location, scale, and content.} 
    \label{fig:varying_conditions}
 \vspace{-20pt}
\end{figure*}

\begin{table}[t]
    \centering
    \caption{Comparison among various sampling manners.}
    \vspace{-10pt}
    \resizebox{\linewidth}{!}{
        \begin{tabular}{c c c c c c}
            \toprule
               & All Sampling &  Random Sampling & $\textbf{topk}(\cdot)$ (Ours) \\ 
            \midrule
            \rowcolor{mygray}
            AP ($\uparrow$) & 24.8 & 21.4 & 22.3 \\
            \rowcolor{mygray}
            T2I-Sim ($\uparrow$)  &  0.3489 & 0.3491 &  0.3513 \\
            
            \bottomrule[1pt] 
    \end{tabular}}
    \label{tab:various_sampling}
    \vspace{-20pt}
\end{table}

\subsection{Visualization Results}

\textbf{Fixing Locations and Scales:} Fig.~\ref{fig:fixed_loc} presents synthesized samples using BoxDiff and samples at each row have the same spatial conditioning input. Given a text prompt ``a \{\} wearing sunglasses'', and the conditioning layout, the location and scale of the animals and sunglasses in the images are consistent with that of the conditional boxes. In addition, one can observe in other rows, the mountains, aurora, castle, and hats are also nearly consistent with the locations and scales of the given conditional boxes. 

\textbf{Visual Comparison:} We present visual comparison between the proposed BoxDiff with the state-of-the-art text-to-image synthesis models such as Stable Diffusion~\cite{sd} and Structure Diffusion~\cite{feng2022training} in Fig.~\ref{fig:compar_and_ablation}. Beyond text prompts as conditional input for image synthesis, additional spatial layout, \emph{e.g.,} box, is used in BoxDiff. One can observe from the figure that, in Stable Diffusion and Structure Diffusion, some subjects are occasionally missing in the synthesized images, \emph{e.g.,} the tie in the second column. Besides, these methods may yield unexpected subjects like the soldier, in which the helmet is actually the target object. In contrast, given spatial conditions, the proposed BoxDiff can correctly synthesize target objects we want in the images. In addition, objects are relatively consistent with the conditional boxes.

\textbf{Varying Locations:} Fig.~\ref{fig:varying_conditions} presents synthesized samples using BoxDiff and samples at each row have one fixed and one varying box constraint. Specifically, given a text prompt of ``a castle in the middle of a calm lake'', the same conditioning inputs are applied for the calm lake, and the conditioning of the castles varies from right to left at the top. Clearly, the lake is always synthesized in the bottom, obeying the given fixed red dashed box. Besides, the location of the castle is changed from the leftmost to the rightmost according to the varying conditioning, \emph{i.e.,} blue dashed box. The same visual variations can also be found in the third row, in which the location of the synthesized castle is moved according to the conditional box. Note that while the text prompts contain the words, \emph{e.g.,} ``in the middle of'', indicating the positional relation between objects, the proposed constraints added on the cross-attentions have a stronger impact on the position of synthetic contents.

\textbf{Varying Scales:} We further probe the controllability of the object scale of the proposed BoxDiff, and visual results are illustrated in the second and third rows of Fig.~\ref{fig:varying_conditions}. Given a text prompt of ``a castle in the middle of a marsh'', we constrain the location and scale of the marsh by a fixed red dashed box and vary the scale of the castle by expanding the red dashed box. Clearly, the size of castles varies from small to large following the given conditional boxes, and the location and scale of the marsh are kept unchanged.

\textbf{Multi-level Variations:} Beyond variation at a single aspect, we simultaneously vary at multiple aspects, \emph{e.g.,} scale, and content, to further demonstrate the effectiveness of our method. As shown in the third row of Fig.~\ref{fig:varying_conditions}, given a text prompt of ``a castle in the middle of \{\}'' and a fixed red dashed box, BoxDiff can successfully synthesize different contents, \emph{i.e.,} ``the grasslands'', ``a sea of sunflowers'', and ``a marsh'', in the red dashed box while controlling the scale of the castle from small to large. 

% These visual results demonstrate that the proposed BoxDiff is capable of simultaneously controlling target objects at different aspects.

\subsection{Quantitative Results}
As shown in Table~\ref{tab:zsp}, we compare fully-supervised layout-to-image methods, \emph{e.g.,} LostGAN~\cite{sun2021learning}, LAMA~\cite{li2021image}, and TwFA~\cite{yang2022modeling} using the newly collected spatial conditions. One can observe from the table, BoxDiff significantly outperforms those fully-supervised ones in terms of the YOLO score. Besides, BoxDiff also achieves the best T2I-Sim, which equivalently represents a better precision of semantic synthesis. Besides, when BoxDiff is integrated, the performance of GLIGEN~\cite{gligen} can be further improved. This validates that our BoxDiff can be used as a plug-and-play component to improve the existing models.

\begin{table}[t]
    \centering
    \caption{Comparison to fully-supervised methods. $^{\dagger}$: model inference with FP16 due to the memory cost.}
    \vspace{-10pt}
    \resizebox{\linewidth}{!}{
        \begin{tabular}{l c c c c c }
            \toprule[1.5pt]
            \multirow{2.5}{*}{Methods} & \multirow{2.5}{*}{Layout Data} & \multirow{2.5}{*}{T2I-Sim ($\uparrow$)}  & \multicolumn{3}{c}{YOLO score} \\
            \cmidrule(lr){4-6}  
            &  &   & AP $(\uparrow)$ & AP$_{50}$ $(\uparrow)$ & AP$_{75}$ $(\uparrow)$ \\
            \midrule
            LostGAN$_{\text{TPAMI'21}}$~\cite{sun2021learning} & COCO-Stuff  & 0.2279 &   5.3 & 8.9 & 5.6  \\
            % Context-L2I & COCO-Stuff  &   \\
            LAMA$_{\text{ICCV'21}}$~\cite{li2021image}&  COCO-Stuff & 0.2396 &   10.2 & 15.3 & 11.7 \\
            TwFA$_{\text{CVPR'22}}$~\cite{yang2022modeling} & COCO-Stuff & 0.2443 & 10.6 & 14.7 & 12.6 \\
            \midrule
            Stable Diffusion~\cite{sd} & None  & 0.3511 & 2.8   & 9.2 & 1.1 \\
            \rowcolor{mygray}
            Stable Diffusion~\cite{sd} + BoxDiff & None  & 0.3513   & 22.3 & 46.8 & 20.2 \\
            \midrule
            GLIGEN$^{\dagger}$~\cite{gligen} & COCO-Stuff  & 0.3489 & 29.7   & 45.8 & 33.9 \\
            \rowcolor{mygray}
            GLIGEN$^{\dagger}$~\cite{gligen} + BoxDiff & COCO-Stuff  & 0.3511 & 40.2   & 62.0 & 46.2 \\
            \bottomrule[1pt] 
    \end{tabular}}
    \label{tab:zsp}
    \vspace{-20pt}
\end{table}

\section{Conclusion and Discussion}
This paper proposed a training-free approach, \emph{i.e.,} BoxDiff, to controlling object synthesis in spatial dimensions. In contrast to conventional layout-to-image methods, the proposed constraints are seamlessly applied to the denoising step of Diffusion models, requiring no additional training. Extensive results demonstrated that BoxDiff enabled the Diffusion models to control objects and contexts where to synthesize. 

To exploit semantic information effectively, we only applied spatial constraints to the cross-attentions at the scale of $16\times 16$. Resolution potentially restricts the precision of the control of object and context synthesis. We believe that as only the simplest form of conditions, \emph{e.g.,} box or scribble, are required, BoxDiff can be potentially extended to data synthesis adhering to additional bounding box conditions, from which a lot of downstream tasks, such as open-vocabulary, weakly- and semi-supervised detection, would benefit. More discussions are included in the appendix.

% \section*{Acknowledgment}
\noindent
\textbf{Acknowledgment}\quad This project is supported by the National Research Foundation, Singapore under its NRFF Award NRF-NRFF13-2021-0008, Mike Zheng Shou's Start-Up Grant from NUS, and the Ministry of Education, Singapore, under the Academic Research Fund Tier 1 (FY2022) Award 22-5406-A0001. Yuexiang Li, Yawen Huang, Haozhe Liu, Wentian Zhang and Yefeng Zheng are funded by Key-Area Research and Development Program of Guangdong Province, China (No. 2018B010111001) and the Scientific and Technical Innovation 2030-"New Generation Artificial Intelligence" Project (No. 2020AAA0104100). Haozhe Liu is also partially supported by the SDAIA-KAUST Center of Excellence in Data Science and Artificial Intelligence (SDAIA-KAUST AI).

{\small
\bibliographystyle{ieee_fullname}
\bibliography{egbib}
}

\clearpage

\section{Appendix}

\subsection{Limitations and Discussion}
\begin{figure}[t]
    \centering
    \includegraphics[width=\linewidth]{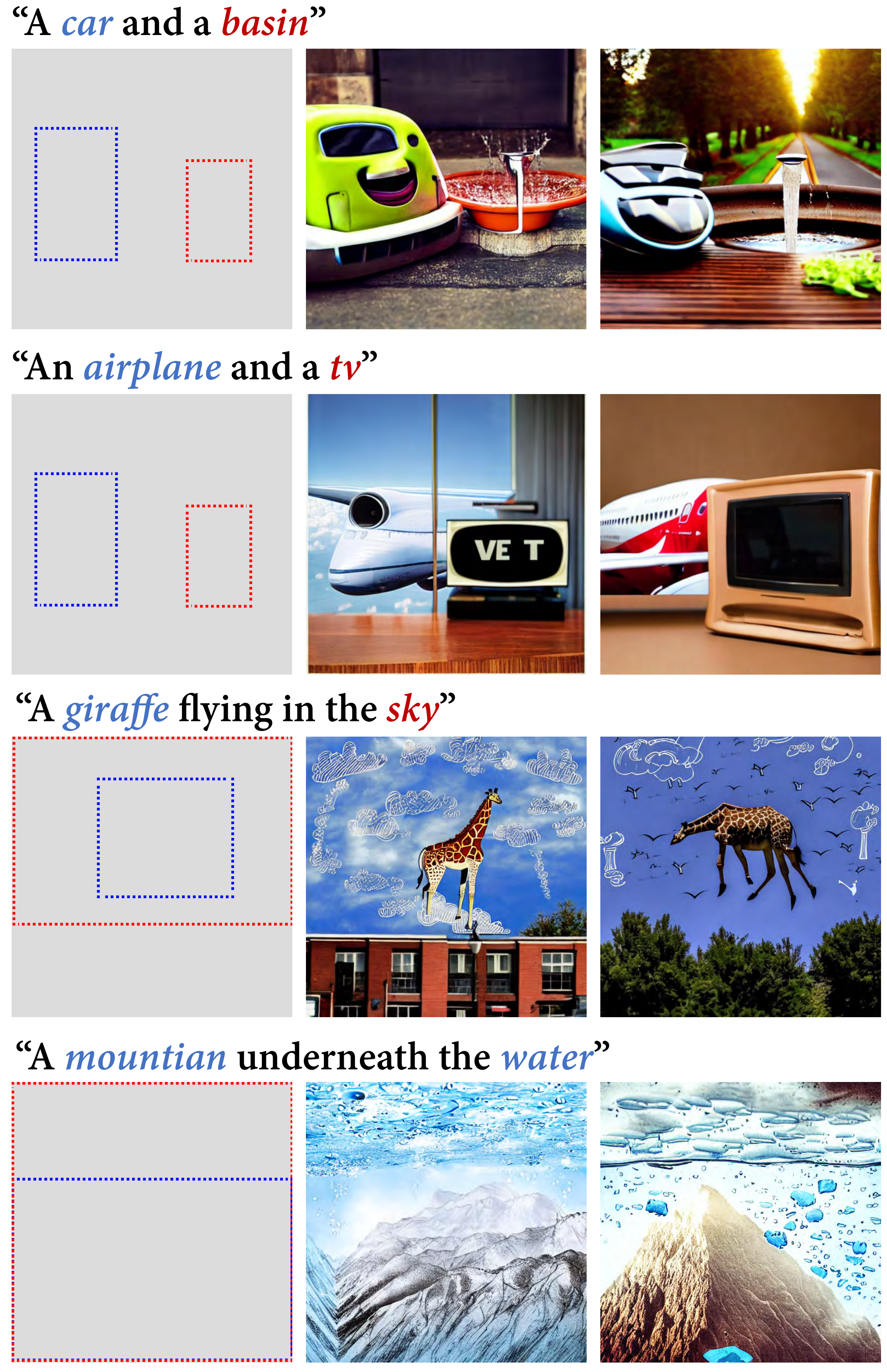}
    \vspace{-20pt}
    \caption{Image synthesis with unusual prompts.} 
    \label{fig:failure_cases}
 \vspace{-20pt}
\end{figure}

We present some synthetic results with unusual prompts in Fig.~\ref{fig:failure_cases}. Here we found some scenarios in which BoxDiff may fail to synthesize realistic images: i) \textbf{combinations of objects that infrequently co-occur}; ii) \textbf{uncommon locations as spatial conditions for objects}. For example in Fig.~\ref{fig:failure_cases}, when ``car'' and ``basin'', which infrequently co-occur, are in a sentence as the text prompt and the uncommon bounding boxes are as the conditions, the synthetic results will be unrealistic and not adhering to the spatial conditions. Besides, the proposed BoxDiff cannot synthesize realistic images when given some uncommon scenes like ``a giraffe flying in the sky'' or ``a mountain underneath the water''.

\begin{figure*}[h]
    \centering
    \includegraphics[width=0.9\linewidth]{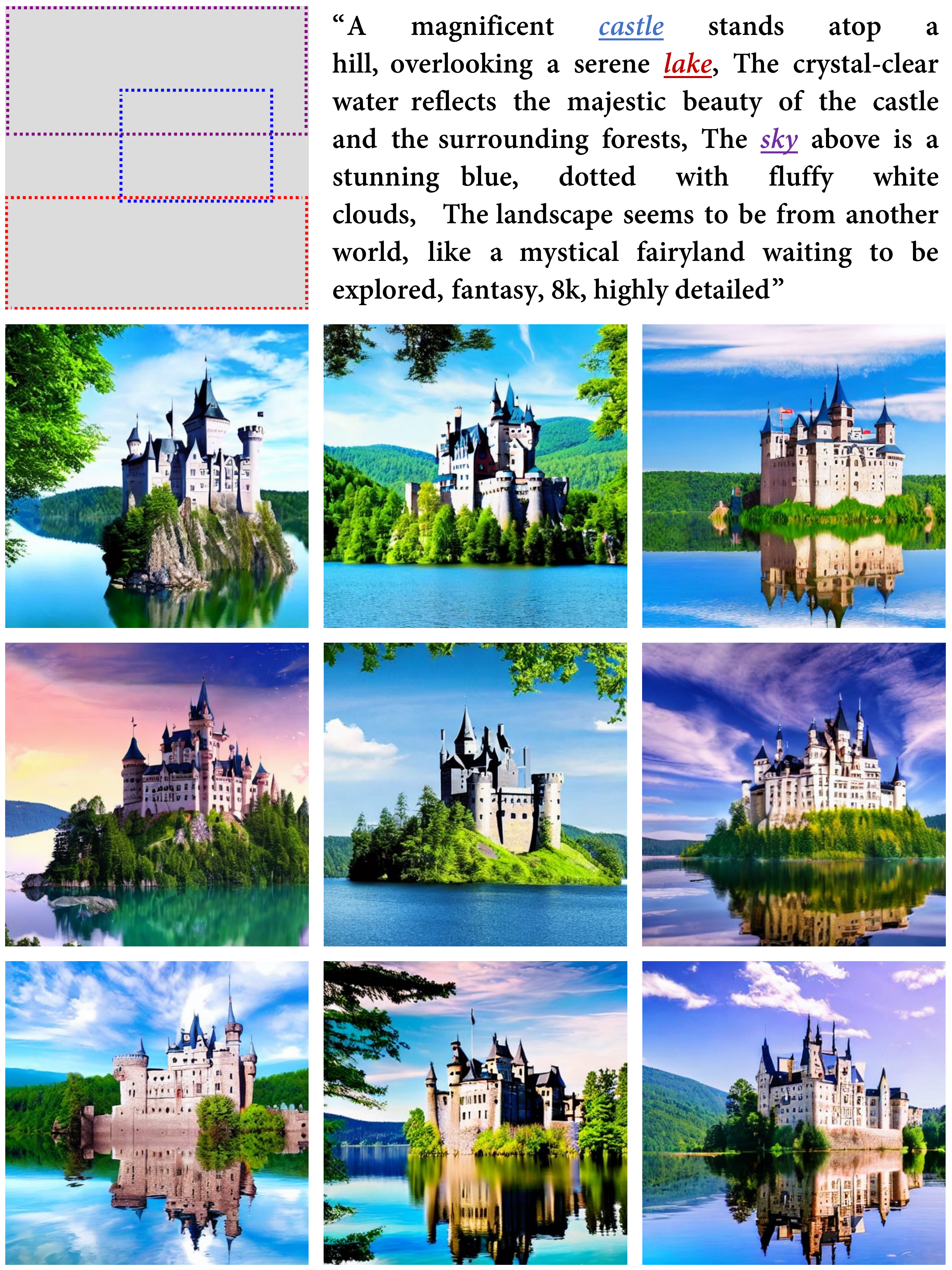}
    \vspace{-12pt}
    \caption{Synthetic castles, lakes, and sky with the same spatial conditions.} 
    \label{fig:vis_more_1}
 % \vspace{-15pt}
\end{figure*}

\begin{figure*}[h]
    \centering
    \includegraphics[width=0.9\linewidth]{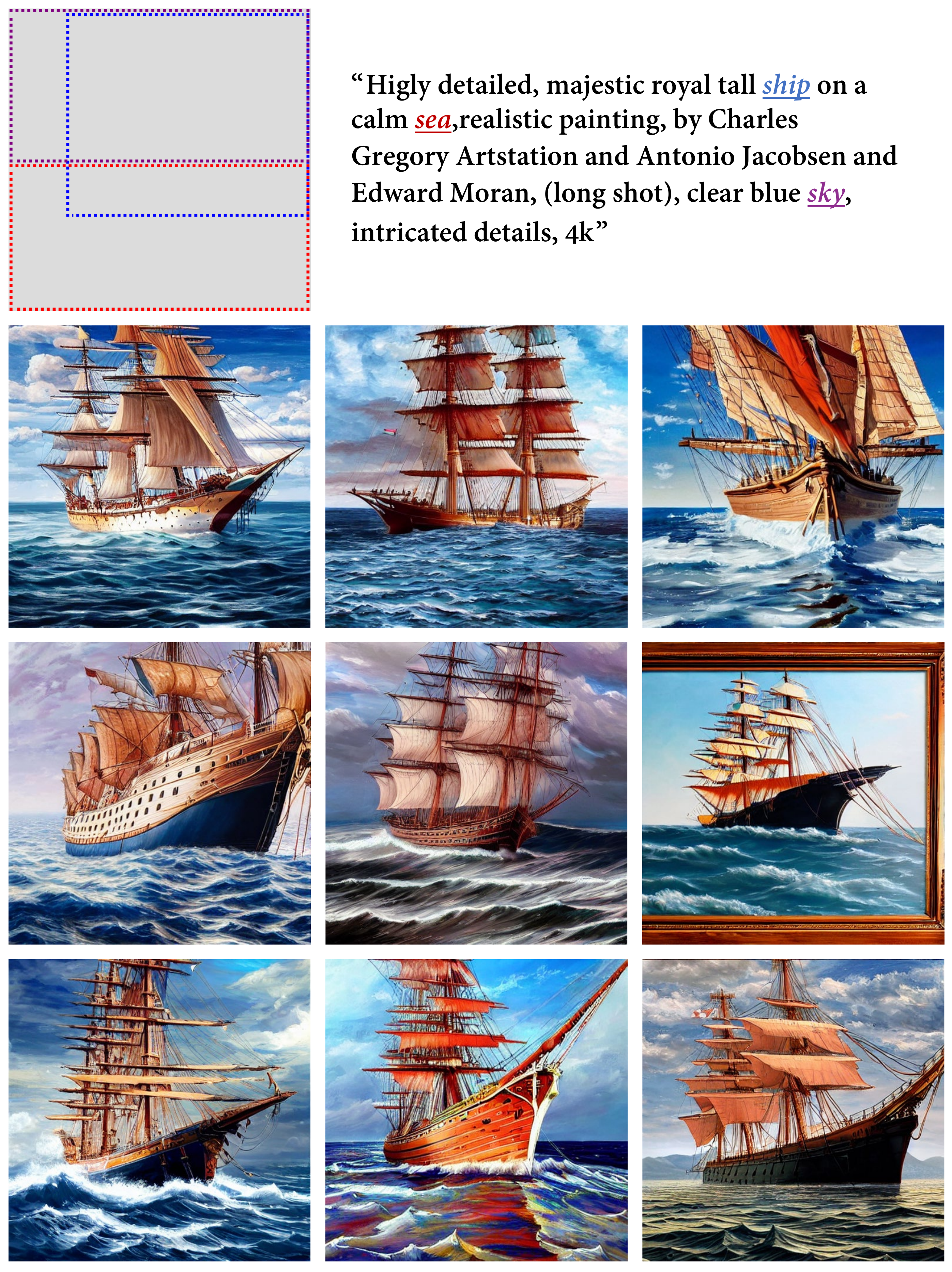}
    \vspace{-12pt}
    \caption{Synthetic ship, sea, and sky with the same spatial conditions.} 
    \label{fig:vis_more_3}
 % \vspace{-15pt}
\end{figure*}

\begin{figure*}[h]
    \centering
    \includegraphics[width=0.9\linewidth]{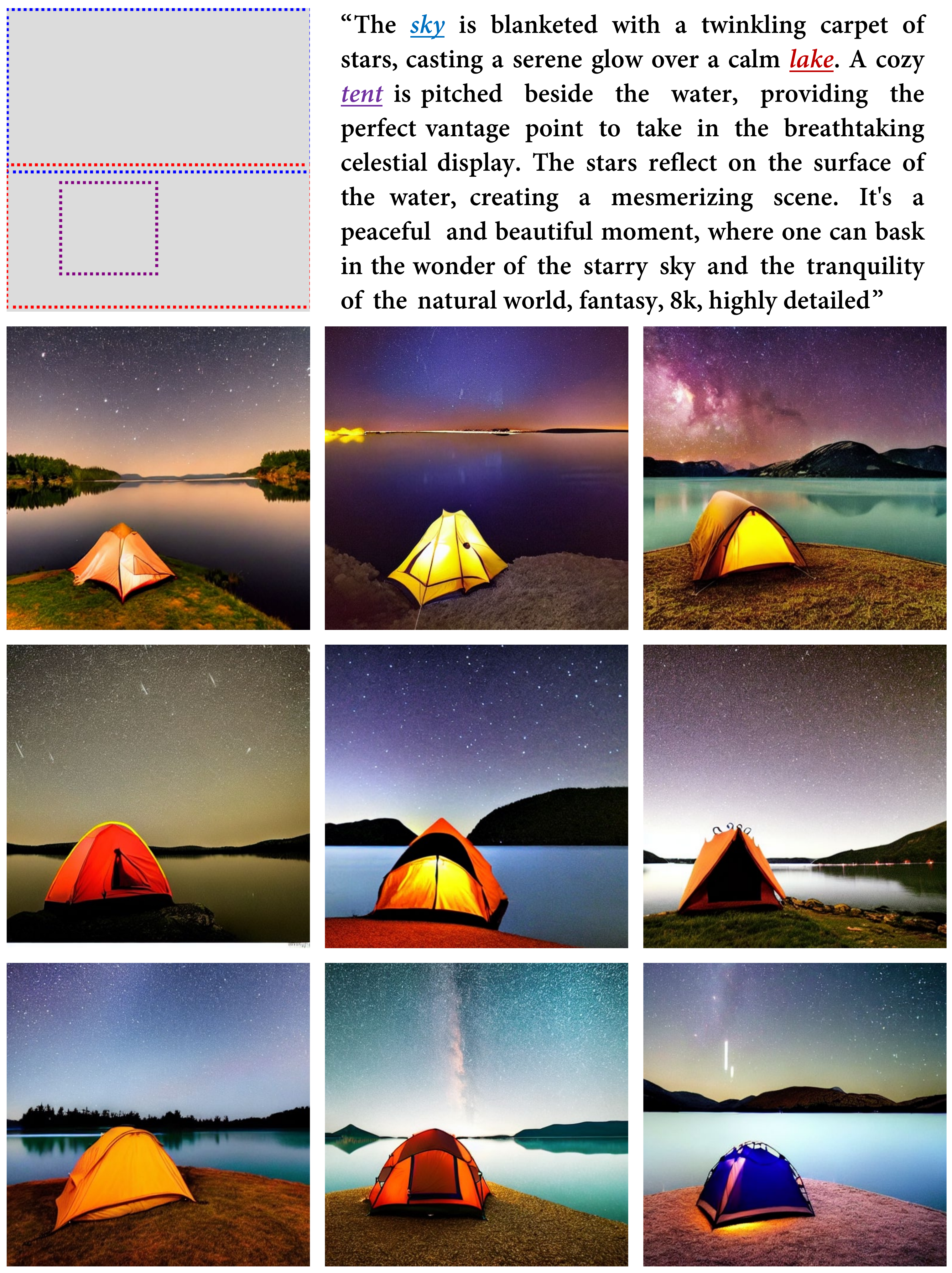}
    \vspace{-12pt}
    \caption{Synthetic starry sky, lakes, and tents with the same spatial conditions.} 
    \label{fig:vis_more_4}
 % \vspace{-15pt}
\end{figure*}

More visualization results are shown in Fig.~\ref{fig:vis_more_1}, \ref{fig:vis_more_3}, and ~\ref{fig:vis_more_4}.

\subsection{Scribble as Conditions}
Simply, scribble can be transformed into bounding boxes, which seamlessly fit the proposed three constraints. In addition, an objectness constraint can be additionally added to further control an object's content or direction. Given a set of scribble $\mathcal{C}=\{\vc_i\}$ with each $\vc_i$ containing Q points $\{(x^i_1,y^i_1),(x^i_2,y^i_2),\cdots,(x^i_Q,y^i_Q)\}$, the objectness constraint can be formulated as:
\begin{equation}
    \mathcal{L}^5_{\mathbf{s}_i}=1 - \frac{1}{Q} \sum \textbf{select}(\mA^t_i, \vc_i),
\end{equation}

\begin{equation}
     \mathcal{L}_{OC} = \sum_{s_i \in \mathcal{S}} \mathcal{L}_{\mathbf{s}_i}^5,
\end{equation}
where $\textbf{select}(\cdot)$ selects corresponding elements using $\vc_i$ from $\mA^t_i$. $\mathcal{L}_{OC}$ can be added to Eq.~(\ref{eq:L}) as the overall constraints when the scribble is given. 

\subsection{Implementation Details}
 All experimental results are obtained using the official Stable Diffusion v1.4 text-to-image synthesis model. The number of denoising steps is set as 50 with a fixed guidance scale of 7.5, and the synthetic images are in a resolution of $512\times 512$. We use a Gaussian kernel with a size of $3\times 3$ and a standard deviation $\sigma = 0.5$. $P$ in $\textbf{topk}(\cdot)$ is set as 80\% of the number of the mask regions $\mM_i$ and $(1-\mM_i)$ so that $P$ is adaptively set according to the size of the mask. $L$ in $\textbf{sample}(\cdot)$ is set as 6, which means that 3 error terms around the given two coordinates are selected, respectively. All experiments are conducted on the NVIDIA TESLA V100 GPU with 32 GB memory.

\begin{figure}[h]
    \centering
    \includegraphics[width=\linewidth]{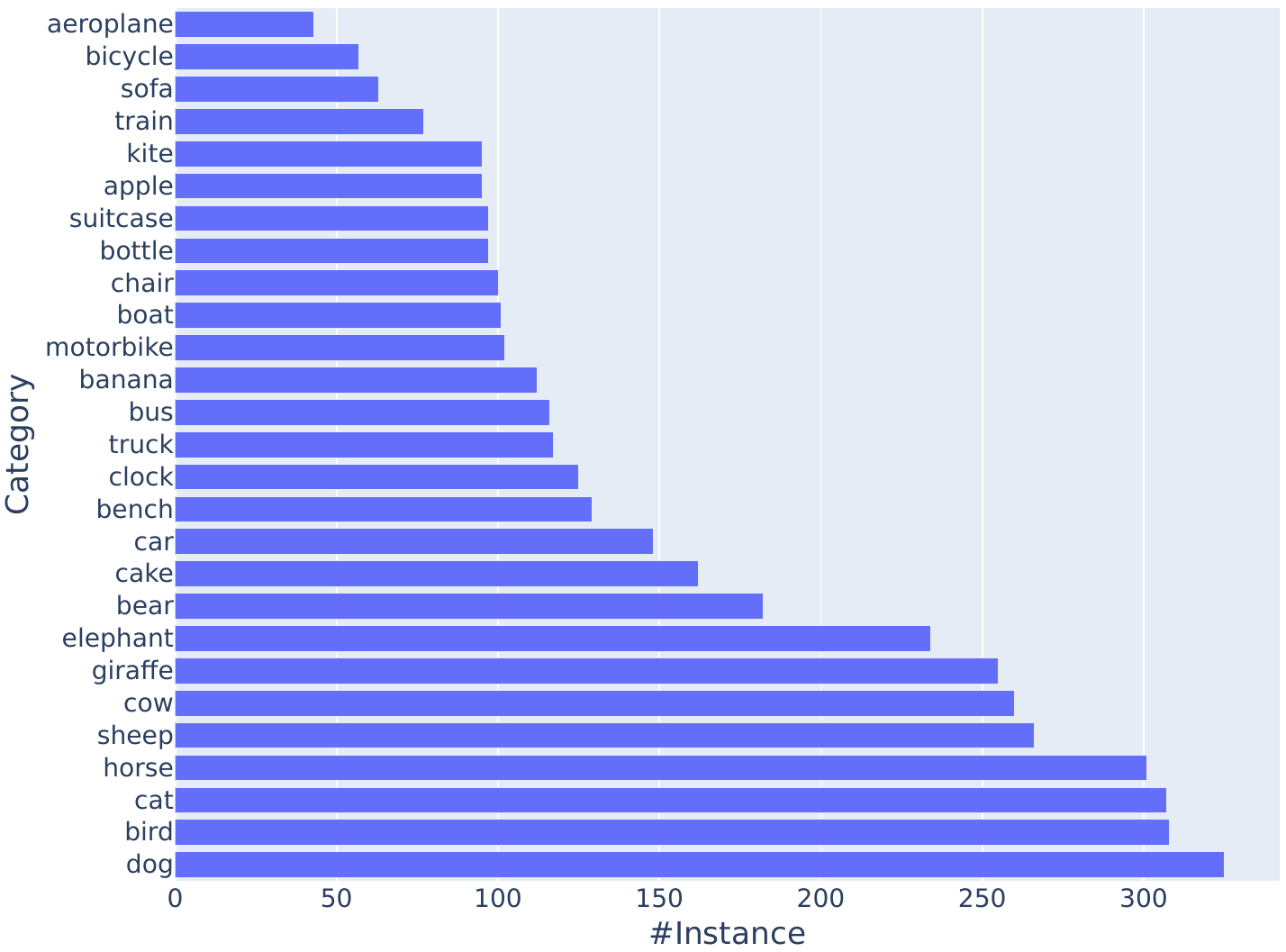}
    \vspace{-20pt}
    \caption{The number of instances of each category in the spatial conditions for zero-shot performance comparison. Here, an instance means a corresponding bounding box.} 
    \vspace{-10pt}
    \label{fig:stat}
\end{figure}

As fully-supervised layout-to-image methods are restricted to a limited scope of categories, we select 9 common animals and 18 common objects from the detection results as the candidate classes. In total, there are 4,274 valid bounding boxes to be the spatial layout conditions. For a fair comparison, we propose to compare the performance of conditional image synthesis on the newly collected layout (no intersection with COCO and VG). The collected candidate categories (9 animals and 18 objects) for spatially conditional text-to-image synthesis are presented below:

 \noindent\fbox{%
    \parbox{0.97\linewidth}{%
    
        \{\, \textcolor{blue}{Animals}: [ 'bird', 'cat', 'dog', 'horse', 'sheep', 'cow', 'elephant', 'bear', 'giraffe'],

          ~~\textcolor{blue}{Objects}: ['bicycle', 'car', 'motorbike', 'aeroplane', 'bus', 'train', 'truck', 'boat', 'bench', 'suitcase', 'kite',
           'bottle', 'banana', 'apple', 'cake', 'chair', 'sofa', 'clock']\}
    }%
}
Fig.~\ref{fig:stat} presents the bar chart for the number of instances of each candidate category. An instance means a corresponding bounding box for conditional image synthesis. It can be seen that there is a maximum number of instances up to 300 and a minimum one around 50. In total, there are 4,274 valid instances (bounding boxes) for image synthesis.
 
\textbf{Selection of Target Tokens:} Typically, given a text prompt such as ``a rabbit and a balloon'', if we are interested in controlling the synthesis of the rabbit and balloon, each single target token or word, e.g., ``rabbit'' and ``balloon'', is enough to extract the corresponding cross-attentions for box-constrained diffusion. However, sometimes we are interested in controlling the objects in the form of compound nouns. For example, given a text prompt such as ``Palm trees sway in the
gentle breeze'', if we aim to control the synthesis of the palm trees, how to perform BoxDiff with two cross-attention maps for a single semantic target? In our experiments, we found that a single token almost dominates the cross-attention for the target semantic. For example, as shown in Fig.~\ref{fig:supp_scribble}, to control the synthesis of palm trees, the cross-attention of ``trees'' is enough for BoxDiff to limit the palm trees within the given conditional box while retaining the correct semantics of ``palm trees''.

\subsection{More Ablation Studies}
In this section, we provide more ablation studies to validate the effectiveness and necessity of $\textbf{topk}(\cdot)$ and $\textbf{sample}(\cdot)$ in Eqs. (4), (6), (10), and (13).

Table~\ref{tab:ablation_topk} presents the influence of various sampling methods on the quality and precision of synthetic images. One can observe from the table that selecting $P$ elements with the largest value in $\textbf{topk}(\cdot)$ obtains the best T2I-Sim and AP. We argue that a pixel within the given box in a higher response represents a higher probability that the object will appear or be synthesized in the pixel. Therefore, sampled elements with a high response obey the prior where the object will appear in the spatial dimension. By contrast, selecting $P$ elements with the smallest value or $P$ random elements in $\textbf{topk}(\cdot)$ achieves a lower T2I-Sim and AP. The comparison further validates the effectiveness and necessity of $\textbf{topk}(\cdot)$ (sampling of $P$ elements with the largest value).

\begin{table}[t]
    \centering
    \caption{Ablation studies on various $\textbf{topk}(\cdot)$.}
    \vspace{-10pt}
    \resizebox{\linewidth}{!}{
        \begin{tabular}{c c c c c c c}
            \toprule[1.5pt]
            
             topk (largest) & topk (smallest) & random & T2I-Sim & AP($\uparrow$)  \\ 
            \midrule
            \rowcolor{mygray}
            \checkmark  &   &   &  0.3513    & 22.3  \\
              & \checkmark  &   & 0.3206  & 12.8 \\
              &  & \checkmark  &   0.3491 & 21.4  \\
            
            \bottomrule[1pt] 
    \end{tabular}}
    \label{tab:ablation_topk}
    \vspace{-10pt}
\end{table}

\begin{table}[t]
    \centering
    \caption{Ablation studies on various $P$ in $\textbf{topk}(\cdot)$.}
        \vspace{-10pt}
    \resizebox{0.6\linewidth}{!}{
        \begin{tabular}{c c c c c c c}
            \toprule[1.5pt]
            
              20\% & 40\% & 80\% & 100\% & T2I-Sim & AP($\uparrow$)  \\ 
            \midrule
            
            \checkmark  &   &   &  & 0.3523  & 13.2  \\
              & \checkmark  &  &  & 0.3516  & 18.5 \\
                \rowcolor{mygray}
              &  & \checkmark  & &  0.3513  & 22.3 \\
            &  & & \checkmark  & 0.3489  & 24.8 \\
            % &  & & \checkmark  & 0.3513  & 22.3 \\
            \bottomrule[1pt] 
    \end{tabular}}
    \label{tab:ablation_p}
    \vspace{-15pt}
\end{table}

In Table~\ref{tab:ablation_p}, we provide the performance of BoxDiff using various $P$ in $\textbf{topk}(\cdot)$ and $P$ is adaptively set according to the percentage of the number of elements in $\mM_i$ and $(1-\mM_i)$. When 80\%  elements of $\mM_i$ and $(1-\mM_i)$ are selected in the Box-Constrained Diffusion, respectively, the BoxDiff can obtain the best T2I-Sim and a relatively higher AP.
\begin{table}[h]
    \centering
 \vspace{-5pt}
    \caption{Ablation studies on various $L$ in $\textbf{sample}(\cdot)$.}
        \vspace{-10pt}
    \resizebox{0.5\linewidth}{!}{
        \begin{tabular}{c c c c c c}
            \toprule[1.5pt]
            
              6 & 10 & 14 & T2I-Sim & AP($\uparrow$)  \\ 
            \midrule
            \rowcolor{mygray}
            \checkmark  &   &    & 0.3513  & 22.3  \\
              & \checkmark  &   & 0.3486  & 22.0 \\
              &  & \checkmark  &   0.3489  & 22.1 \\
            % &  & & \checkmark  & 0.3472  & 7.7 \\
            % &  & & \checkmark  & 0.3513  & 22.3 \\
            \bottomrule[1pt] 
    \end{tabular}}
    \label{tab:ablation_sample}
    \vspace{-15pt}
\end{table}

We conduct an ablation study to determine the value of $L$ used in Eqs.~(10) and (13) and the results are presented in Table~\ref{tab:ablation_sample}. When $L=6$, the best T2I-Sim and AP are achieved. When $L$ varies from 6 to 14, though AP is relatively stable, T2I-Sim is accordingly decreased. This validates that more constraints on the cross-attentions will affect the quality of the synthetic images, and representative sampling is sufficient for image synthesis obeying the spatial conditions while retaining higher image quality.

\subsection{More Visualization Results}
\begin{figure*}[h]
    \centering
    \includegraphics[width=0.9\linewidth]{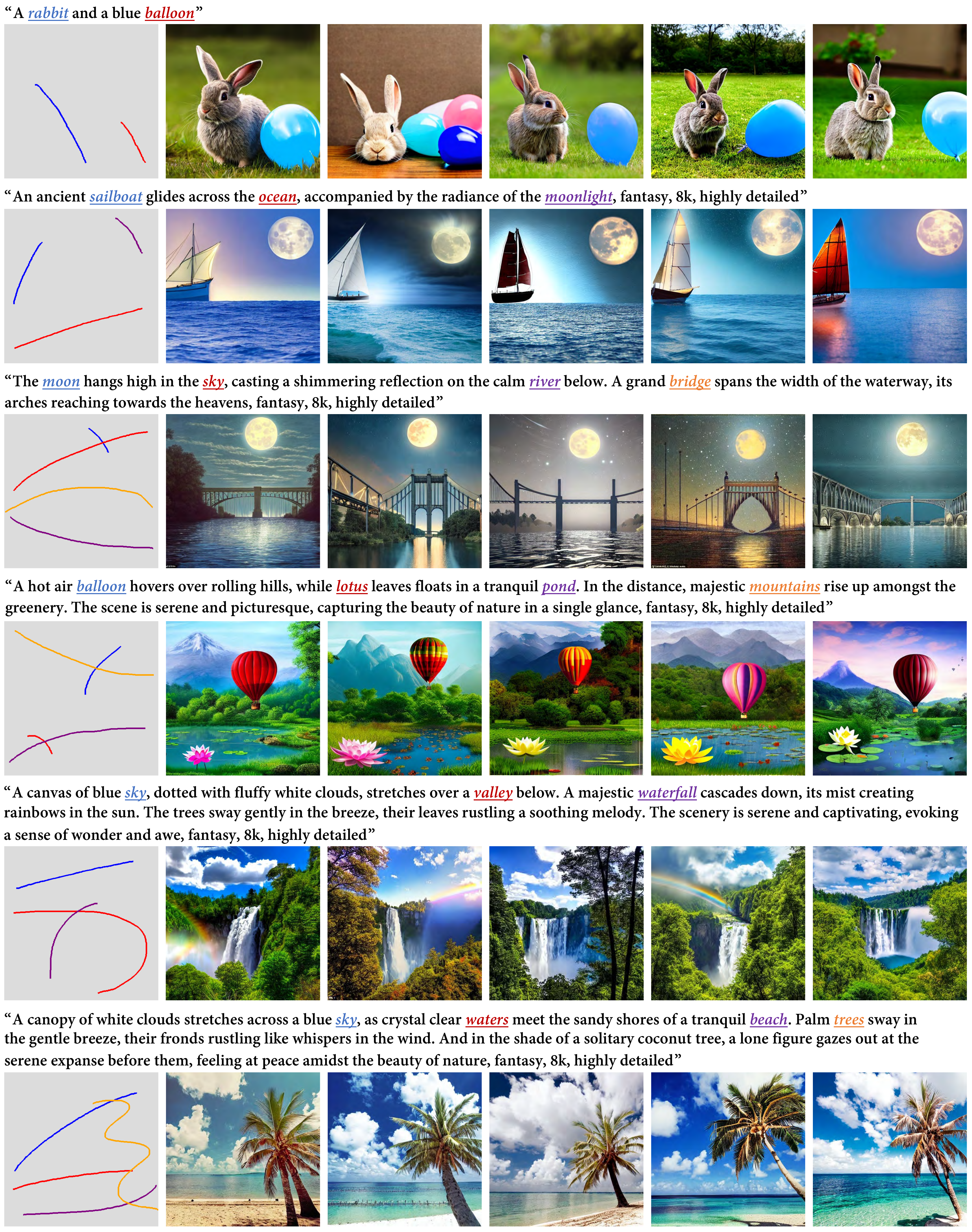}
    \vspace{-12pt}
    \caption{Synthetic samples using scribble spatial conditions.} 
    \label{fig:supp_scribble}
 % \vspace{-15pt}
\end{figure*}

\begin{figure*}[h]
    \centering
    \includegraphics[width=0.9\linewidth]{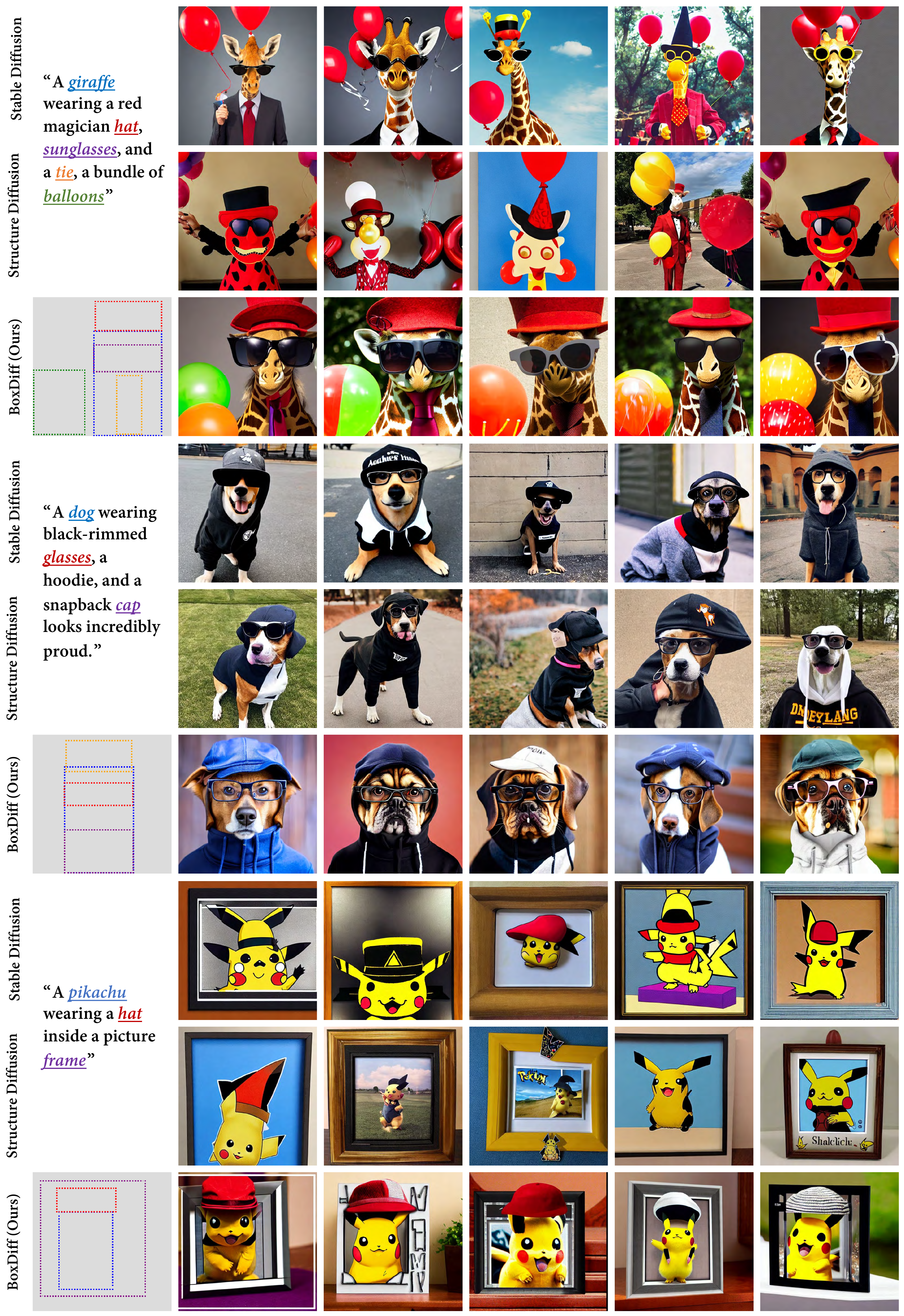}
    \vspace{-12pt}
    \caption{More visual comparison among Stable Diffusion, Structure Diffusion, and the proposed BoxDiff.} 
    \label{fig:supp_visual_compar}
 % \vspace{-15pt}
\end{figure*}

\textbf{Scribble as Conditions:} As discussed in the paper, the proposed BoxDiff can also interact with other types of spatial conditions such as scribble. Here, we provide more synthetic samples using scribble conditions in Fig.~\ref{fig:supp_scribble}. Beyond object-bounding boxes, scribble provides more pixel information about the object content. This motivates the proposed objectness constraints $\mathcal{L}_{OC}$, which can further control the object's content or direction. For example, as shown in the second row of Fig.~\ref{fig:supp_scribble}, the content of the sailboat and palm trees are relatively consistent with the scribble conditions, \emph{i.e.,} the blue and orange line. 
\textbf{More Visual Comparisons:} In Fig.~\ref{fig:supp_visual_compar}, we provide more visual comparison among Stable Diffusion, Structure Diffusion, and the proposed BoxDiff. It can be seen that contents such as tie and hat occasionally are missed in the samples synthesized by Stable Diffusion and Structure Diffusion. In contrast, samples generated by the proposed BoxDiff are relatively consistent with the spatial conditions. Besides, each target object is correctly presented in the resulting images.  

%%%%%%%%% ABSTRACT
\begin{figure*}[h]
    \centering
    \includegraphics[width=0.9\linewidth]{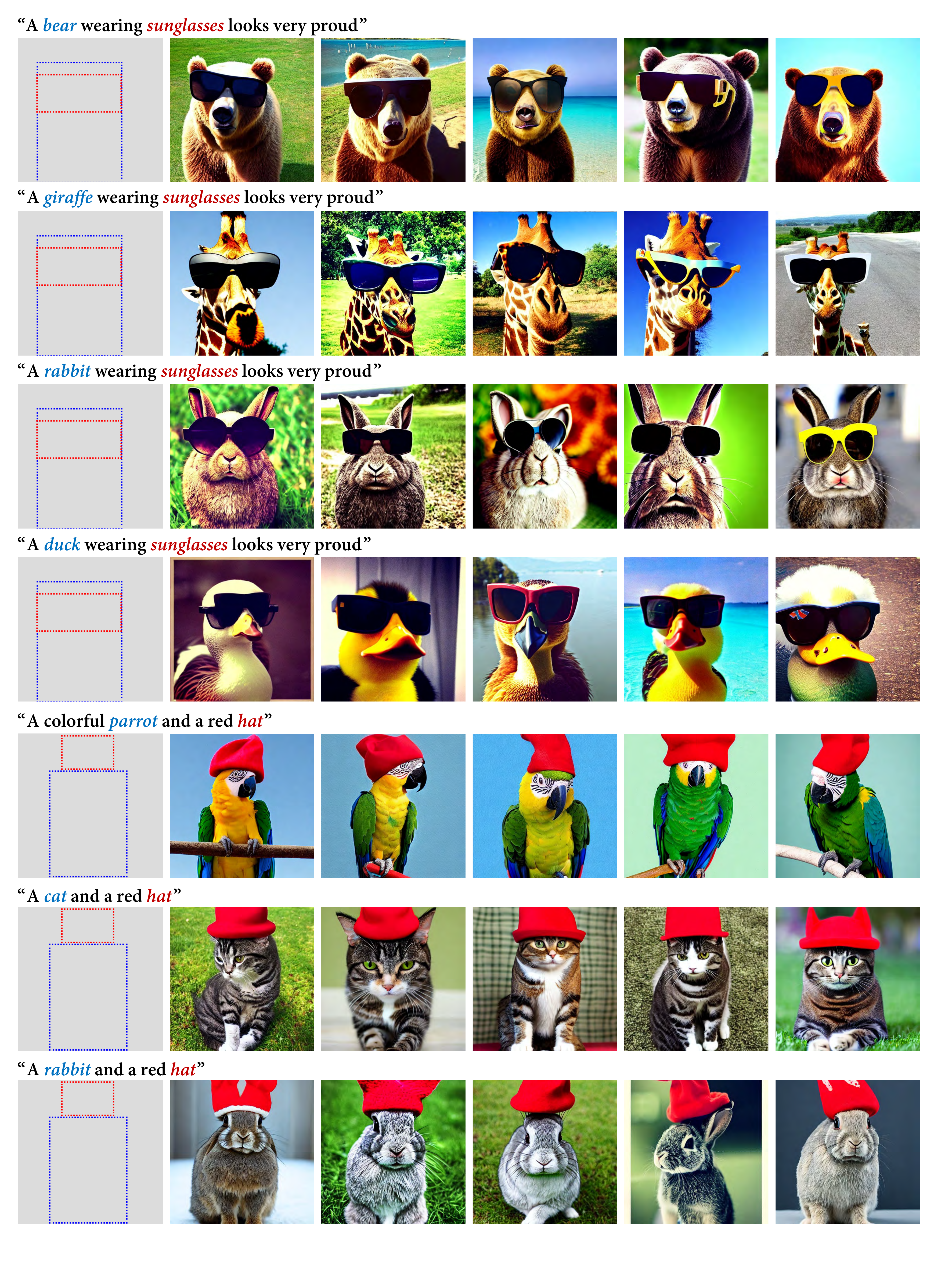}
    \vspace{-12pt}
    \caption{Synthetic funny animals wearing sunglasses or a red hat.} 
    \label{fig:funny_animials_sunglasses_hat}
 % \vspace{-15pt}
\end{figure*}

\begin{figure*}[h]
    \centering
    \includegraphics[width=0.9\linewidth]{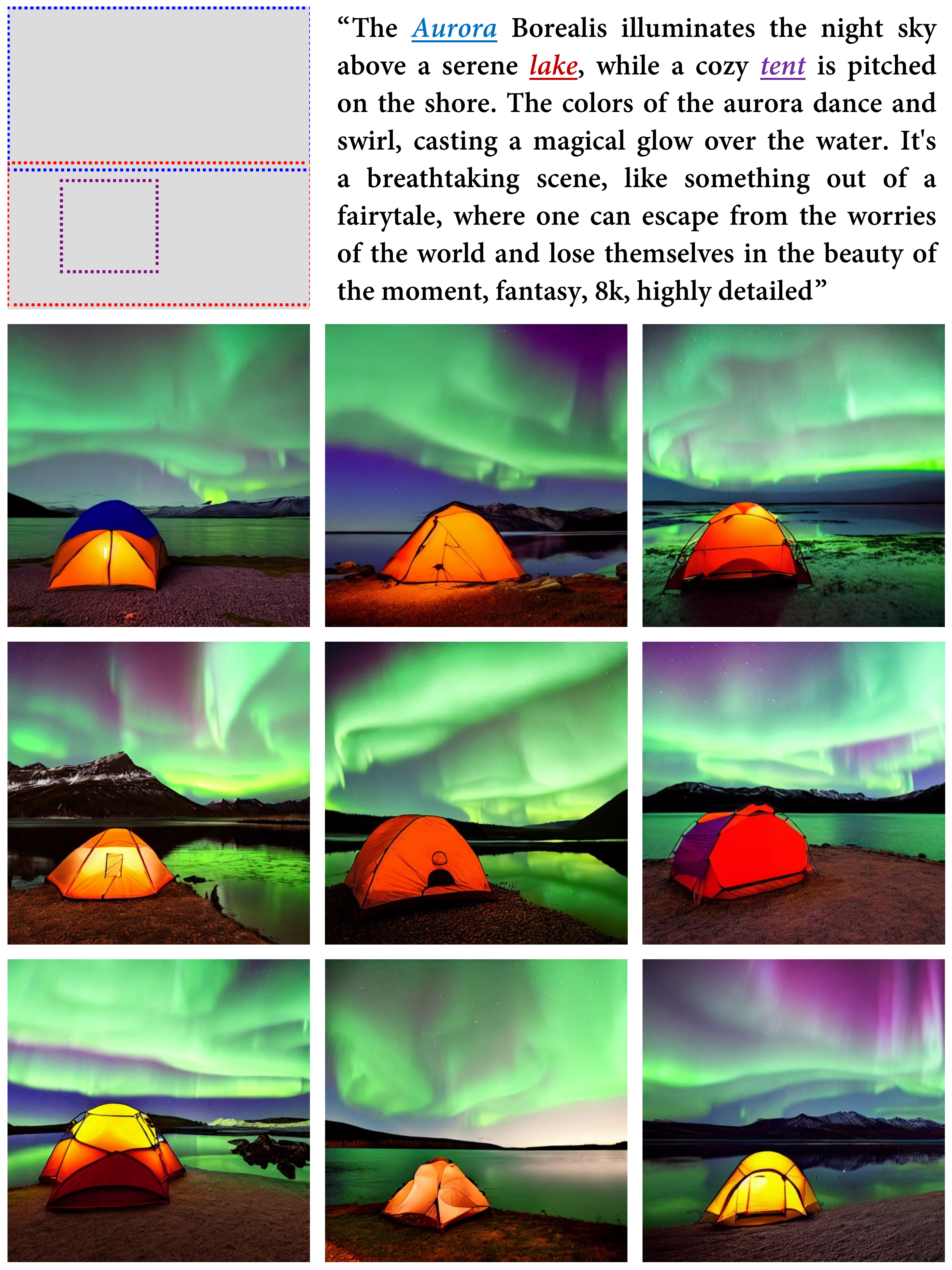}
    \vspace{-12pt}
    \caption{Synthetic aurora, lakes, and tents with the same spatial conditions.} 
    \label{fig:vis_more_2}
 % \vspace{-15pt}
\end{figure*}

\end{document}